\documentclass{article}

\usepackage{microtype}
\usepackage{graphicx}
\usepackage{subfigure}
\usepackage{booktabs} 
\usepackage{multirow}

\usepackage{hyperref}

\usepackage[accepted]{icml2020}

 \newcommand{\bx}{\mathbf{x}}

\icmltitlerunning{Explainable Deep Modeling of Tabular Data using TableGraphNet}

\begin{document}

\twocolumn[
\icmltitle{Explainable Deep Modeling of Tabular Data using TableGraphNet}



\icmlsetsymbol{equal}{*}

\begin{icmlauthorlist}
\icmlauthor{Gabriel Terejanu}{uncc}
\icmlauthor{Jawad Chowdhury}{uncc}
\icmlauthor{Rezaur Rashid}{uncc}
\icmlauthor{Asif Chowdhury}{uofsc}
\end{icmlauthorlist}

\icmlaffiliation{uncc}{Department of Computer Science, University of North Carolina at Charlotte, Charlotte, North Carolina, USA}
\icmlaffiliation{uofsc}{Department of Computer Science and Engineering, University of South Carolina, Columbia, South Carolina, USA}

\icmlcorrespondingauthor{Gabriel Terejanu}{gabriel.terejanu@uncc.edu}

\icmlkeywords{interpretability, feature attribution, graph networks, importance scores}

\vskip 0.3in
]



\printAffiliationsAndNotice{}  

\begin{abstract}
The vast majority of research on explainability focuses on post-explainability rather than explainable modeling. Namely, an explanation model is derived to explain a complex black box model built with the sole purpose of achieving the highest performance possible. In part, this trend might be driven by the misconception that there is a trade-off between explainability and accuracy. Furthermore, the consequential work on Shapely values, grounded in game theory, has also contributed to a new wave of post-explainability research on better approximations for various machine learning models, including deep learning models. We propose a new architecture that inherently produces explainable predictions in the form of additive feature attributions. Our approach learns a graph representation for each record in the dataset. Attribute centric features are then derived from the graph and fed into a contribution deep set model to produce the final predictions. We show that our explainable model attains the same level of performance as black box models. Finally, we provide an augmented model training approach that leverages the missingness property and yields high levels of consistency (as required for the Shapely values) without loss of accuracy.
\end{abstract}

\section{Introduction}
\label{introduction}

The interest in explaining model predictions is increasing. In part, this is due to regulations in industries such as insurance to ensure that models do not discriminate. It is also driven by users trying to extract actionable insights moving beyond observations and predictions to influence the business process.

Most of the research efforts are spent on post-explainability rather than explainable modeling. Post-explainability approximates the original model with an explanation model. However, this approach creates an artificial trade-off between accuracy and explainability. The best explanation should be provided by the model itself~\cite{rudin2019}. As a result, we believe that explainability should be embedded into the model to allow the model to make the right predictions for the right reasons.

We propose a general architecture, called TableGraphNet, to build predictive models that also provide feature attribution during prediction. The focus of the architecture is on providing local interpretability~\cite{DoshiKim2017Interpretability}, where the goal is to explain a particular prediction of a model and an input instance, as compared with global interpretability that provides an overall behavior of the model. The explainability provided by TableGraphNet is in the form of feature attribution, where the output of the model is decomposed in the contribution of each attribute~\cite{Shrikumar2017}. We show that the architecture does not suffer from an accuracy vs explainability trade-off. We actually observe, that better explainability yields better accuracy.

In recent years we have seen a number of post-explanatory attribution methods~\cite{Ribeiro2016,kindermans2017learning,zintgraf2017visualizing,Montavon_2017,Sundararajan2017,NIPS2017_7062,pmlr-v97-ancona19a}. Unfortunately, a number of these approaches provide unreliable contributions~\cite{Kindermans2019}, prompting a shift in the research direction to adopt an axiomatic view of the problem~\cite{NIPS2017_7062,Sundararajan2017,pmlr-v97-ancona19a}. This research direction has its origins in Shapely values~\cite{Shapley1953}. However, computing Shapely values is an intractable problem in general and requires various approximations to speed up their calculation~\cite{NIPS2017_7062}. 

We recognize the importance of the properties that Shapely values need to have, not only for post-explainability approaches but also for explainable modeling like in this case. Adopting an explainable model does not guarantee explainability. Without an axiomatic framework, assessing the quality of attributions/contributions, even when produced by an explainable model, is a daunting task. It is also difficult to assess the quality empirically, as there is a lack of benchmark datasets labeled with the true importance of the attributes. The design of the TableGraphNet incorporates two of the properties required for Shapely values (local accuracy and missingness~\cite{NIPS2017_7062}) and we empirically observe that consistency~\cite{NIPS2017_7062} can also be met by adopting an augmented training strategy that exploits the missingness property. 

Enforcing these properties yields additional benefits, such as the fact that TableGraphNet naturally deals with missing values in the original dataset and it does not require imputation. Furthermore, TableGraphNet also obeys the symmetry-preserving property~\cite{pmlr-v97-ancona19a}, namely the output is not impacted by the order of the attributes.

In the context of explainable modeling in computer vision,~\cite{chen2018looks} has proposed to add a prototype layer to conventional convolutional neural networks (CNN) to capture semantic concepts in image classification. Modifications to CNNs are also proposed by~\cite{Zhang_2018} by associating each higher level filter with an object part. Joint prediction and explanation of object recognition is also proposed by~\cite{Hendricks_2016}. For natural language processing,~\cite{Lei_2016} propose to extract pieces of text as justifications for predictions using a modular architecture containing a generator and encoder.

An augmented training dataset that contains an explanation along with the original attributes is proposed by~\cite{Hind_2019}. A similar augmented dataset approach is provided by~\cite{Park_2018}. Unfortunately, both approaches assume the existence of explanations in the training dataset.

Finally, regularization can also be used to improve explainability. For example,~\cite{wu2017sparsity} regularizes models such that their decision boundary can be approximated by decision trees, which are inherently explainable. A different approach is proposed by~\cite{Ross2017} to penalize input gradients using domain expert annotations.

TableGraphNet targets tabular data, which is intrinsically heterogeneous, and most of the time, each attribute defines a clear concept (e.g. customer credit score). It adopts the axiomatic view proposed for Shapely values and does not require any annotations in the data. In the process, we believe that this architecture also fills a gap on the applicability of neural networks to tabular data. In this context, we have not seen neural networks having the same impact as compared with their stellar success in computer vision~\cite{lecun-98,10.1145/3065386}, speech~\cite{WaveNet}, and natural language processing~\cite{BERT}, where data is homogeneous and has spatial and temporal dimensions.

Section~\ref{tablegraphnet} introduces the architecture proposed for TableGraphNet. Section~\ref{experiments} details the numerical experiments and the results obtained using TableGraphNet. Finally, Section~\ref{conclusions} summarizes the study and provides future directions for improvement. 

\section{TableGraphNet}
\label{tablegraphnet}

Given $M$ attributes for each record $\bx = [x_1, \ldots x_M]$ in the dataset, the goal is to develop a predictive function $f(\bx)$ that is decomposable in $M$ terms corresponding to the individual contribution, $\rho_k(\bx)$, of each attribute $x_k$, for $k=1 \ldots M$, and $\rho_0$, a biased term independent of the data $\bx$.
\begin{equation}
f(\bx) = \rho_0 + \sum_{k=1}^M \rho_k(\bx) \label{decomposition}
\end{equation}

Note that $f(\bx)$ is not an explanation model that approximates a more complex black box model. Instead, it is the main model used to make predictions, and as a byproduct it provides also information regarding feature attribution.

\subsection{Architecture}

Even architectures that follow the decomposition in Eq.\ref{decomposition} may yield confounding contributions. To address the problem of designing an architecture that incorporates an additive feature attribution, we have used as inspiration the work of J\"{o}rg Behler and Michele Parrinello~\cite{Behler2007,behlersymfun} on determining the molecular atomization energy using additive subnetwork models and atom-centric features. In this context, the molecule is represented as a chemical graph where vertices correspond to the atoms and edges correspond to chemical bonds. Various symmetry functions are proposed to extract atom centric features. An example of a symmetry function is one that aggregates the pairwise distances information centered around each atom. Finally, the atomic centric features of each atom are processed by an atomic neural network which is shared across all the atoms. The final energy is just the sum of all the individual atomic energies calculated using the corresponding atom neural network.

The architecture proposed in~\cite{Behler2007,behlersymfun} is an additive deep set model~\cite{NIPS2017_6931} where the inputs are atom centric features and it provides invariance with respect to atom permutation and molecular size extensibility. We propose to extend this type of architecture to accommodate tabular data by overcoming the challenge that tabular data contains heterogeneous attributes and it does not have a natural graph representation to be able to extract attribute centric features.

\begin{figure*}[h]
\vskip 0.2in
\begin{center}
\centerline{\includegraphics[width=\linewidth]{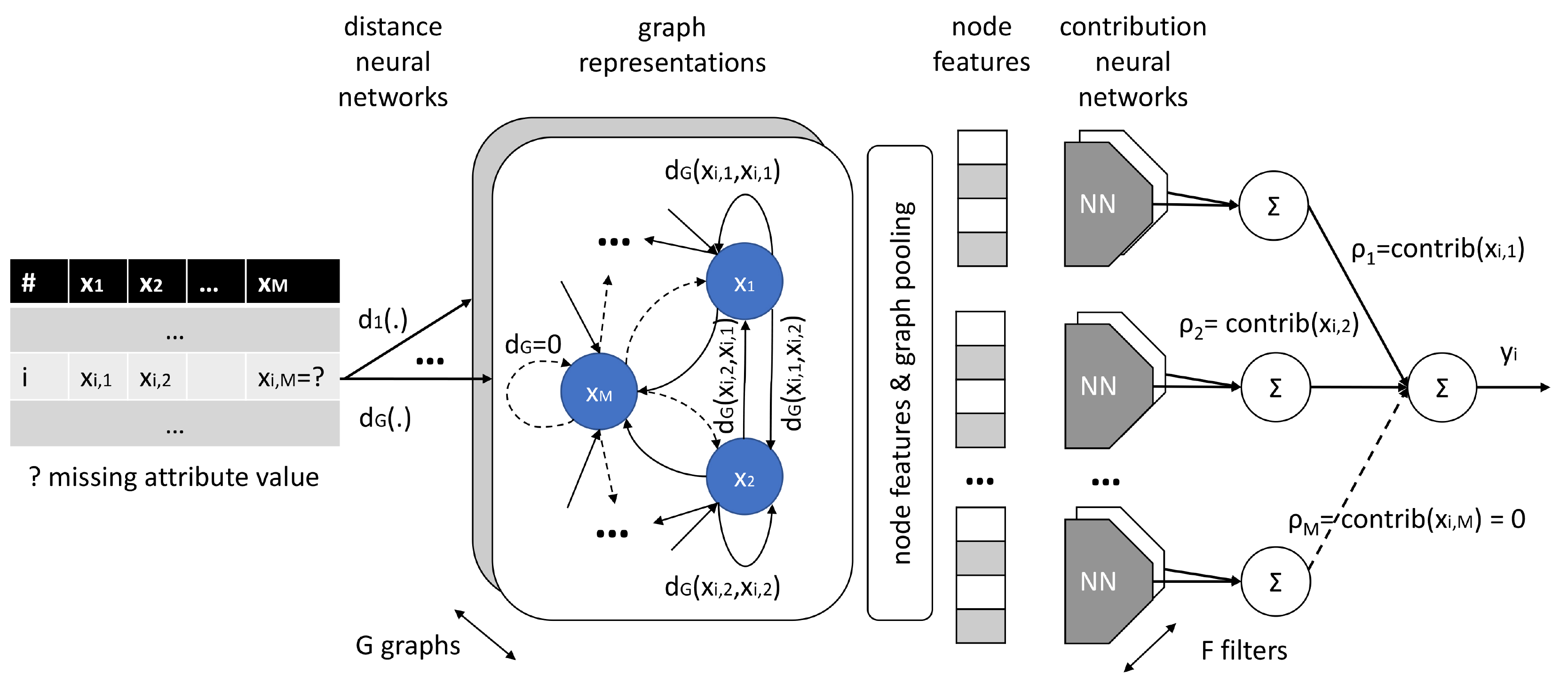}}
\caption{TableGraphNet overall architecture. Each record in the dataset is represented as a set of graphs induced by distance neural networks learned during model training. From each graph, node centric features are extracted and pooled across all the graphs to form attribute centric features. These are fed into a set of contribution neural networks (jointly learned with the distance neural networks) to produce the final attribute contributions, the sum of which yields the model prediction that approximates the desired target $y$.}
\label{overall}
\end{center}
\vskip -0.2in
\end{figure*}

The overall architecture of the TableGraphNet is depicted in Fig.~\ref{overall}. The idea flows naturally from the previously cited work. Each record/row in the dataset is represented as a graph or a series of graphs. Within a graph, the vertices correspond to original attributes and edges correspond to distances between these attributes. With a graph representation of each record, we can then extract attribute centric features (e.g. an aggregation of the pairwise distances centered around each attribute), which finally become inputs into an additive deep set model that predicts the target. The terms in the final addition represent the contributions of each individual attribute. The graph representations are obtained using a set of distance neural networks which are trained at the same time with the contribution neural networks used in the deep set model. 

In the followings we detail the components of the proposed architecture: (1) initial data preparation, (2) cartesian product transformation, (2) graph representations using distance neural networks, (3) attribute centric features and pooling, and (4) contribution neural networks.

\textbf{Initial data preparation} includes flattening the dataset such that we obtain an $N \times M$ table, where $N$ is the number of records and $M$ is the number of attributes. Since we plan to learn a distance function between two attributes, all the attributes are scaled to the same range.

\begin{figure*}[h]
\vskip 0.2in
\begin{center}
\centerline{\includegraphics[width=\linewidth]{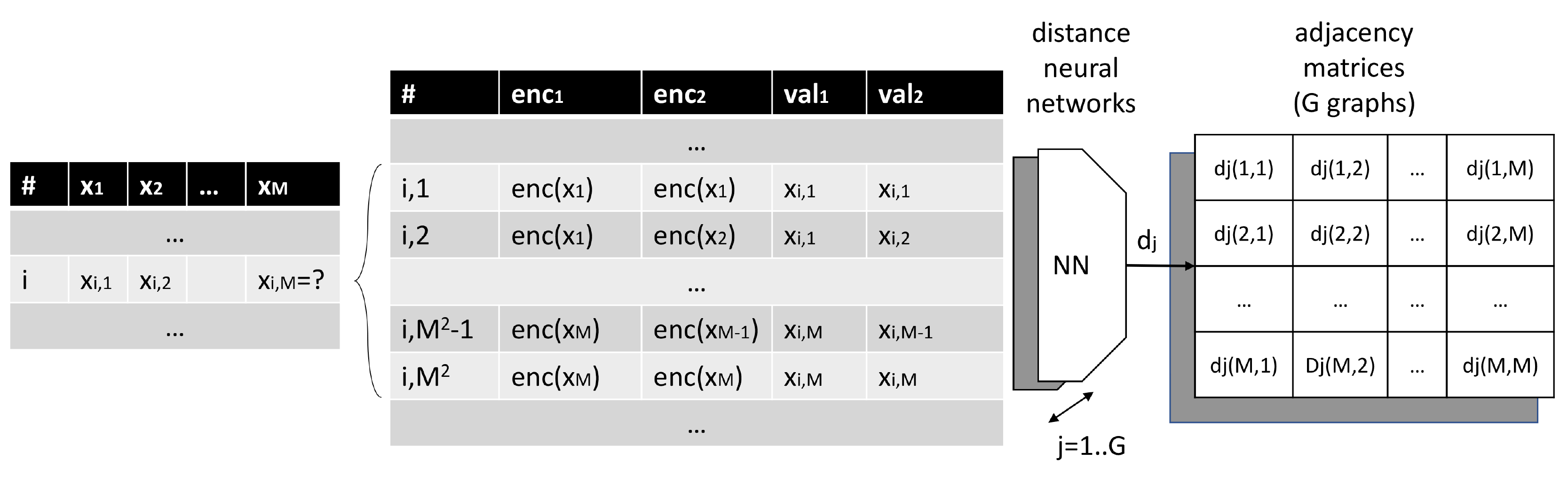}}
\caption{Cartesian product transformation and distance neural networks. Each record of $M$ attributes is transformed into $M^2$ pairwise attribute data used in distance calculations. Every attribute is represented using its value as well as an encoding, such as one-hot-encoding, for attribute identification and better model performance. We assume that any attribute is connected with all the other attributes and the distance that defines the edge value is learned during training. Every distance neural network is shared across all the pairs of attributes and we use multiple distances to improve the performance of the model.}
\label{cartesian}
\end{center}
\vskip -0.2in
\end{figure*}

\textbf{Cartesian product transformation} is performed to turn the $N \times M$ dataset into an $NM^2 \times 2(E+1)$ dataset, where $E \ge 0$, see Fig.~\ref{cartesian}. Since for each record we need to calculate the distance between all the pairs of attributes, this results in $M^2$ distance calculations for each record. Each distance calculation has at least two arguments namely the two values corresponding to the two attributes involved in the calculation. 

In addition to the two attribute values, one may choose to also add the encodings corresponding to the two attributes (e.g. their one-hot-encodings, where $E$ represents the number of encoding features). We have noticed numerically that including the one-hot-encodings of the two attributes significantly improves the accuracy of the model. The inclusion of an attribute identification introduces an additional degree of freedom and allows the model to differentiate between various pairs of attributes. For example, the distance calculation between the credit score and credit limit of a customer (positive correlation) should be different than the distance between the credit score and interest rate (negative correlation).

\textbf{Graph representations using distance neural networks.} Each record/row in the dataset is represented using a set of graphs $\mathcal{G} = \{ G_j \}_{j=1\ldots G}$, where each graph is defined as a tuple $G_j = \{ V_j, E_j \}$. The $V_j = \{x_k\}_{k=1 \ldots M}$ is the set of nodes corresponding to the original attributes, and the $E_j = \{ (e_{j,k,l}, x_k, x_l) \}_{k,l=1\ldots M}$ is the set of edges, where $x_k$ is the source, $x_l$ is the receiver, and $e_{j,k,l}$ is the distance from $x_k$ to $x_l$ in the $j$th graph. 

The distance between nodes is modeled using a neural network, parameters of which are learned during the training of the TableGraphNet. The same distance function is shared across all the pairs of attributes.
\begin{equation}
e_{j,k,l} = d_j^{\textrm{NN}}(\textrm{enc}(k), \textrm{enc}(l), x_k, x_l)
\end{equation}

Here, $\textrm{enc}(k)$ represents the encoding of the $k$th attribute and $x_k$ is the value of the $k$th attribute for a specific record in the dataset. The record number has been omitted in the above formula to improve readability, but the formula should be considered in the context of the $i$th record.

Note that this distance function is the one that induces a graph representation for each record in the dataset. While one can choose to use a predefine distance (e.g. L1 or L$2$ norm) without attempting to learn it as suggested here, we have found that jointly learning distances along with the contribution neural networks significantly improves the accuracy of the model.

Furthermore, distances such as L1 or L$2$ norms are proper symmetric distances that obey the triangle inequality. We do not impose any constraints in learning the distance function $d_j^{\textrm{NN}}$. We believe that to obtain a good representation for attribute contributions one requires asymmetric distance functions.
\begin{equation}
e_{j,k,l} \ne e_{j,l,k}
\end{equation}

This assertion is not backed by any theoretical result at this point. We only have anecdotal evidence from numerical results, and the intuition that asymmetric edges will result in more diverse node centric features used in attribute contribution calculations. The assumption is that similar node centric features will result in confounding attribute contributions.

\textbf{Attribute centric features and pooling.} Given that each record in the dataset is now represented using a set of graphs induced by the learned distance functions, we can proceed to extract node/attribute centric features from each graph and then pool them across the graphs. Example of features include the aggregation (e.g. sum, mean, max) of the values of outflow and/or inflow edges for each node in the graph, or the node rank determined using PageRank~\cite{BRIN1998107}. Consider the sum of values of outflow edges.
\begin{equation}
a_{j,k} = \sum_{l=1}^M e_{j,k,l}
\end{equation}

The final attribute centric features are the result of aggregating the corresponding node centric features across all the graphs (e.g. concatenation, max). Consider the concatenation operation.
\begin{equation}
\mathbf{a}_k = [ a_{j,k} ]_{j=1\ldots G}
\end{equation}

Note that in the current architecture and numerical experiments we have incorporated all the information from the original attributes into the values of the edges, and did not added additional information at the node level, which is the reason why all the node centric features previously introduced are derived from edges. Because of this limitation, we found numerically that to achieve the lowest predictive error, we need more than one graph representation.

Nevertheless, the model can be extended by using graph networks~\cite{GraphNets} to derive new graphs from the original induced graphs based on additional parameterized operations for information propagation over nodes and edges. The information from these derived graphs can be significantly richer than currently used in our study, and will be explored in future studies.

\textbf{Contribution Neural Networks.} The attribute centric features extracted and pooled over all the graph representations, $\mathbf{a}_k$, are fed into a set of contribution neural networks to calculate the individual contribution of each attribute. A contribution neural network is shared across all the feature groups corresponding to the original attributes. In addition, to improve the performance of the model, we 
propose to treat the shared weights of contribution neural networks as a filter in a convolutional neural network~\cite{Chowdhury2020}.

We use $F$ filters to learn different features similarly with basic image features such as edges or corners~\cite{cnnrep} that are subsequently combined to detect higher level concepts such as faces~\cite{alexnet}. Each filter in the set, scans the corresponding graph based features of each original attribute, and to obtain the contribution of one attribute, $\rho_k$, we sum the outputs of all the filters for that particular attribute, $\rho_{k,j}$. 
\begin{eqnarray}
\rho_{k,j} &=& c_j^{\textrm{NN}}(\mathbf{a}_k), ~\textrm{for}~j=1\ldots F \\
\rho_k &=& \sum_{j=1}^F \rho_{k,j}
\end{eqnarray}

The final output of TableGraphNet is given by summing all the attribute contributions along with a bias term $\rho_0$ that is learned during model training.
\begin{equation}
f(\bx) = \rho_0 + \sum_{k=1}^M \rho_k
\end{equation}

As previously mentioned, this part of the model is just an additive deep set model~\cite{NIPS2017_6931}, which ensures that if the relative order of the original attributes is changed, it will not impact the final output of the model. 

\subsection{Relation to Shapely values}
\label{sec:consistency}

In this section, we use the axiomatic set introduced by~\cite{NIPS2017_7062} to uniquely determine the additive feature attributions. The first property, \textbf{local accuracy}, it is inherently satisfied by our proposed additive model, Eq.~\ref{decomposition}. Note that $\rho_0 = f(\mathbf{0})$, where $f(\mathbf{0})$ is the value of the function when all the attribute values are missing and it is learned during model training.

The second property, \textbf{missingness}, states that the contribution of the $k$th attribute, $\rho_k = 0$ if the value of this attribute is missing. In TableGraphNet this is enforced by (1) setting all the pairwise features to zero if the value of the source attribute is missing, (2) using distance neural networks and contribution neural networks without bias terms, such that the zeros get propagated through the network, and by (3) judiciously choosing the node features (e.g. sum of values of outflow edges). Note that in this case, all the node centric features for the missing attributes are zero.

The third property, \textbf{consistency}, states that if the value of the model changes due to an increase in the importance of an attribute regardless of the rest then the contribution of that attribute should not decrease. For this particular property, we do not have any theoretical result to show that TableGraphNet guarantees consistency, however, numerically we find that choices in the node centric features as well as how the model is trained do have an impact on consistency. 

Additionally, because TableGraphNet implements an additive deep set, it also obeys the \textbf{symmetry preserving property} as proposed by~\cite{NIPS2017_7062} and~\cite{pmlr-v97-ancona19a}. Namely the output of the model is independent on the order of the attributes, and the attributions or contributions follow the same order of the attributes.

\subsection{Augmented training strategy}
\label{sec:training}

By exploiting the missingness property, we have devised an augmented training strategy that generates an infinite dataset from a table by randomly choosing records to be used in training and attributes to be marked as missing. Numerically we have seen that as the number of epochs increases the percentage of the consistency conditions that we have checked converges towards full consistency. More exploration remains to be done to further understand the role of missingness during training. For now, we have only tested marking an attribute as missing in a record with $50\%$ probability.

Interestingly, trying to enforce missingness and consistency yields another benefit: a natural approach to deal with missing data. As a result, TableGraphNet, does not require imputation for the missing values in the original dataset. 

\section{Experiments}
\label{experiments}

All numerical experiments ran on a $4\times$NVIDIA V100 32GB computational server using a TensorFlow~\cite{tensorflow2015-whitepaper} implementation.

\subsection{Performance Comparison}

To evaluate the performance of TableGraphNet, we have used $8$ datasets for regression and $3$ datasets for classification from the UCI repository~\cite{Dua:2019}. The performance of TableGraphNet is compared with the performance of a dense neural network, as well as with the performance of a multifilter neural network, which instead of learning the distance for graph representations is using a predefined distance.

The performance metric for regression is the average root mean squared error (RMSE) along with its standard deviation obtained over $5$ trial runs. For classification, we have also used $5$ trial runs to calculate the average and standard deviation of area under the curve (AUC) with macro averaging for multiclass problems. For each trial run, we have randomly shuffled the dataset and used $80\%$ of the data for training and validation and $20\%$ of the data for testing.

All the models have been trained for at most $10,000$ epochs with an early stopping criteria based on the validation loss (mean squared error for regression, and cross entropy for classification). The validation data is based on $20\%$ of the original $80\%$ of data used for training and validation. All models have been optimized using the Adam optimizer~\cite{kingma2015adam}. In the following paragraphs we detail how we have obtained the three models used in the performance comparison.

\begin{table*}[h]
\centering
\begin{tabular}{|l|l|l|l|l|l|}
\hline
\textbf{Datasets}   & \textbf{N} & \textbf{M} & \textbf{Dense NN} & \textbf{TableGraphNet} & \textbf{Multifilter NN} \\ \hline
Boston Housing      & 506        & 13         & 3.34 $\pm$ 0.28       & 3.29 $\pm$ 0.50            & 4.52 $\pm$ 0.34             \\ \hline
Concrete Strength   & 1030       & 8          & 5.89 $\pm$ 0.27       & 4.93 $\pm$ 0.18            & 8.48 $\pm$ 0.33             \\ \hline
Energy Efficiency   & 768        & 8          & 1.12 $\pm$ 0.09       & 1.12 $\pm$ 0.12            & 2.73 $\pm$ 0.28             \\ \hline
Kin8nm              & 8192       & 8          & 2.16 $\pm$ 0.13       & 2.02 $\pm$ 0.04            & 2.14 $\pm$ 0.02             \\ \hline
Naval Propulsion    & 11934      & 16         & 0.01 $\pm$ 0.00       & 0.25 $\pm$ 0.00            & 0.01 $\pm$ 0.00             \\ \hline
Power Plant         & 9568       & 4          & 4.16 $\pm$ 0.19       & 3.99 $\pm$ 0.11            & 4.78 $\pm$ 0.08             \\ \hline
Wine Quality Red    & 1599       & 11         & 0.63 $\pm$ 0.03       & 0.69 $\pm$ 0.03            & 0.78 $\pm$ 0.05             \\ \hline
Yacht Hydrodynamics & 308        & 6          & 4.61 $\pm$ 1.16       & 1.00 $\pm$ 0.21            & 8.85 $\pm$ 0.83             \\ \hline
\end{tabular}
\caption{Regression results on UCI datasets. The performance metric is the average and standard derivation of test RMSE (root mean squared error) obtained using $5$ trial runs. Here, $N$ is the number of records in the dataset and $M$ is the number of attributes. With the exception of Naval Propulsion, the RMSE of TableGraphNet is better or statistically not different from the Dense NN. The subpar performance of the Multifilter NN is an indication that distance learning is beneficial.}
\label{tab:regression}
\end{table*}

\textbf{Regression: TableGraphNet.} To determine an optimum architecture for TableGraphNet for regression, we have used a grid search strategy on the Boston Housing dataset and chose the best $4$ performers and used only these models on all the other datasets and reported the best performance of the four. We have varied the number of graphs: $\{1, 8\}$, the number of hidden layers and neurons for the distance neural networks has been kept fixed at $(16,8)$, we have used the sum of outflow and inflow edges as node-centric features and concatenated them over all the graphs. The number of contribution filters has been varied: $\{1,8\}$, and the architecture of the contribution neural network was chosen from the following set $\{ (16), (24, 8), (64, 16, 4) \}$. We have used a kernel L$2$ regularization with coefficients $\{0.01, 0.5\}$ and the following learning rates $\{ 0.0001, 0.001\}$. The TableGraphNet has been trained using an early stopping with a minimum delta of $0.01$ over a patience period of $200$. 

\textbf{Regression: Dense NN.} For reference, we have used a dense neural network with the following hidden layers and number of neurons: $\{ (16), (24, 8), (64, 16, 4) \}$. As with the TableGraphNet, we have used a kernel L$2$ regularization with coefficients $\{0.01, 0.5\}$ and the following learning rates $\{ 0.0001, 0.001\}$. The Dense NN has been trained using an early stopping with a minimum delta of $0.01$ over a patience period of $200$. The grid search for Dense NN has been performed on the Boston Housing dataset and the top $6$ performers have been used across the rest of the datasets and we are reporting the best performance of the six.

\textbf{Regression: Multifilter NN.} A variant of the TableGraphNet has been also tested. The difference is that the distance function is fixed, namely the absolute value between the values of two attributes. As a result we have just one induced graph. To compensate for the lack of additional graphs as compared with TableGraphNet we have used a number of node centric features such as the original value of the attribute, node2vec~\cite{node2vec} (dimension: $5$, walk length: $13$, number of walks: $3$), and betweenness centrality~\cite{freeman1977set}. These node centric features have been used with contribution filters of size $\{4,8,12\}$ with the following architectures: $\{ (16), (6,6), (7,8), (24,8), (64,16,4) \}$. The learning rates considered for Adam optimizer are $\{0.1, 0.01, 0.001\}$ and the coefficients for the L$2$ regularizations are $\{0.1, 0.001\}$. The Multifilter NN has been trained using an early stopping with a minimum delta of $0.001$ over a patience period of $500$. As compared with TableGraphNet and Dense NN, the grid search to obtain the optimum Multifilter NN model has been performed for each individual dataset.

\textbf{Regression results} are presented in Table~\ref{tab:regression}. We note that with the exception of Naval Propulsion, the RMSE of TableGraphNet is better or statistically not different from the Dense NN, while the Multifilter NN has an overall worse performance than both TableGraphNet and Dense NN.

\begin{table*}[h]
\centering
\begin{tabular}{|l|l|l|l|l|l|}
\hline
\textbf{Datasets}  & \textbf{N} & \textbf{M} & \textbf{Dense NN} & \textbf{TableGraphNet} & \textbf{Multifilter NN} \\ \hline
Kin8nm             & 8192       & 8          & 0.85 $\pm$ 0.01       & 0.86 $\pm$ 0.00            & 0.51 $\pm$ 0.02             \\ \hline
Wine Quality Red   & 1599       & 11         & 0.82 $\pm$ 0.02       & 0.83 $\pm$ 0.00            & 0.64 $\pm$ 0.00             \\ \hline
Wine Quality White & 4898       & 11         & 0.75 $\pm$ 0.02       & 0.69 $\pm$ 0.01            & 0.58 $\pm$ 0.03             \\ \hline
\end{tabular}
\caption{Classification results on UCI datasets. The performance metric is the average and standard derivation of test AUC (area under the curve using macro averaging) obtained using $5$ trial runs. Here, $N$ is the number of records in the dataset and $M$ is the number of attributes. With the exception of Wine Quality White, the AUC of TableGraphNet is  statistically not different from the Dense NN. As with the regression experiments, we partially attribute the subpar  performance of the Multifilter NN to its pre-defined distance calculation.}
\label{tab:classification}
\end{table*}

\textbf{Classification - TableGraphNet.} For classification we have used the followings: number of graphs: $\{2,16\}$, distance neural network architecture: $(16,8)$, number of filters: $\{2,16\}$, filter architecture: $\{(16), (32,16,4), (64,32,16,4)\}$, L$2$ regularization coefficients: $\{0.01, 0.1, 0.5\}$, learning rates: $\{0.001, 0.0001\}$, minimum delta of $0.01$ and patience $200$ for early stopping. A separate grid search has been performed for each classification dataset.

\textbf{Classification - Dense NN.} The followings have been used in the grid search for the Dense NN: architecture: $\{(16), (32,16,4), (64,32,16,4)\}$, L$2$ regularization coefficients: $\{0.01, 0.1, 0.5\}$, learning rates: $\{0.001, 0.0001\}$, minimum delta of $0.01$ and patience $200$ for early stopping. A separate grid search has been performed for each classification dataset.

\textbf{Classification - Multifilter NN.} The same node centric features as in regression have been used also for classification. The following settings have been used in the grid search for the Multifilter NN: number of filters: $\{4, 12\}$, filter architecture: $\{(16), (32,16,4)\}$, L$2$ regularization coefficients: $\{0.001, 0.1\}$, learning rates: $\{0.01, 0.001, 0.0001\}$, minimum delta of $0.001$ and patience $500$ for early stopping. A separate grid search has been performed for each classification dataset.

\textbf{Classification results} are presented in Table~\ref{tab:classification}. With the exception of Wine Quality White, the AUC of TableGraphNet is statistically not different from the Dense NN. As with the regression results, the Multifilter NN performs worse than both TableGraphNet and Dense NN.

Overall, we find the results promising in that an explainable architecture does not have to trade accuracy for explainability. We also observe that learning the distance between the attributes improves the performance over using a predefined distance as in the Multifilter NN, even though it uses much richer node centric features. Note that features such as node2vec, since are not differentiable are challenging to be incorporated into an architecture such as TableGraphNet, which is trained using a gradient based approach.

\subsection{Consistency Check}

Assessing the quality of attributions/contributions is a daunting task as it requires a labeled dataset to include the importance scores for all the attributes. In this section, we are concerned with checking the consistency property as discussed in Section~\ref{sec:consistency}. The consistency property~\cite{NIPS2017_7062} states that given any two models $f()$ and $f'()$ if $f'(\mathbf{z}) - f'(\mathbf{z}/k) \ge f(\mathbf{z}) - f(\mathbf{z}/i)$ for all inputs $\mathbf{z}$ derived from $\bx$ having any subset of missing attributes then $\rho_k'(\bx) \ge \rho_i(\bx)$. Here, the notation $\mathbf{z}/k$ is equivalent with the input $\mathbf{z}$ having the $k$th attribute marked as missing.

For this task we have performed the experiments on the UCI handwritten digits dataset~\cite{Dua:2019}. Note that TableGraphNet is recommended to be used for attributes that represents clear defined concepts rather than pixels. Furthermore, because it needs to create a fully connected graph it is not suitable (at least not our current implementation) for thousands of attributes but rather at most hundreds.

However, since this dataset has only $64$ attributes, we have checked the consistency condition for all $k=1\ldots 64$ attributes and all the samples in the testing dataset  that is $30\%$ of the original dataset. Since this is a multiclass problem, we have set $f(), f'()$ to the output of the model before the softmax layer and alternate the functions between various digit outputs (e.g. $f()=f_5()$ and $f'()=f_9()$). This has resulted in $3,104,640$ conditions to be checked and we have reported the percentage of conditions when the consistency is met. In addition to consistency, we also report accuracy. Both performance measures have been averaged over $5$ trial runs. The difference between the trial runs is just the random initial weights of the models.

\begin{table*}[h]
\centering
\begin{tabular}{|l|c|l|c|l|}
\hline
\multicolumn{1}{|c|}{\multirow{2}{*}{\textbf{Model}}}   & \multicolumn{2}{c|}{\multirow{2}{*}{\textbf{Accuracy}}} & \multicolumn{2}{c|}{\multirow{2}{*}{\textbf{Consistency}}} \\
\multicolumn{1}{|c|}{}                                  & \multicolumn{2}{c|}{}                                   & \multicolumn{2}{c|}{}                                      \\ \hline
Classic training; 100 epochs; sum(outflow), sum(inflow) & \multicolumn{2}{c|}{0.927$\pm$0.008}                        & \multicolumn{2}{c|}{0.657$\pm$0.026}                           \\ \hline
Classic training; 100 epochs; sum(outflow)              & \multicolumn{2}{c|}{0.943$\pm$0.017}                        & \multicolumn{2}{c|}{0.860$\pm$0.018}                           \\ \hline
Augmented training; 100 epochs; sum(outflow)            & \multicolumn{2}{c|}{0.927$\pm$0.006}                        & \multicolumn{2}{c|}{0.911$\pm$0.011}                           \\ \hline
Augmented training; 500 epochs; sum(outflow)            & \multicolumn{2}{c|}{0.931$\pm$0.011}                        & \multicolumn{2}{c|}{0.961$\pm$0.010}                           \\ \hline
Augmented training; 1000 epochs; sum(outflow)           & \multicolumn{2}{c|}{0.944$\pm$0.010}                        & \multicolumn{2}{c|}{0.964$\pm$0.009}                           \\ \hline
\end{tabular}
\caption{Consistency results using the UCI handwritten digits dataset.  The performance metrics are averaged over 5 trial runs. We observe that augmented training for a larger number of epochs increases both the accuracy and the consistency of the contributions.}
\label{tab:consistency}
\end{table*}

Table~\ref{tab:consistency} reports the performance measures for $5$ types of models used. All the TableGraphNet models have the same architecture. Namely, $32$ graphs with the distance architecture $(32,16)$, $8$ filters, with the contribution architecture $(32,16)$, L$2$ regularization coefficient of $0.02$, and a learning rate of $0.001$ that has an exponential decay rate of $0.96$. We consider the model with classic training strategy, where the records do not have any missing data, as well as the proposed augmented training where a record has random attributes marked as missing as discussed in Section~\ref{sec:training}. 

The first observation is that the type of node centric features impacts both accuracy as well as consistency. The second observation is that by using the augmented training strategy we see an increase in the consistency percentage. This increase persists as the number of epochs is increased and seems to be positively correlated with the accuracy of the model. Namely, better explainability yields better accuracy.

We also note that we do not see a significant degradation in the accuracy by switching to the augmented training strategy, but it does require more epochs due to data missingness. 

\begin{figure}[h]
\vskip 0.2in
\begin{center}
\centerline{\includegraphics[width=\linewidth]{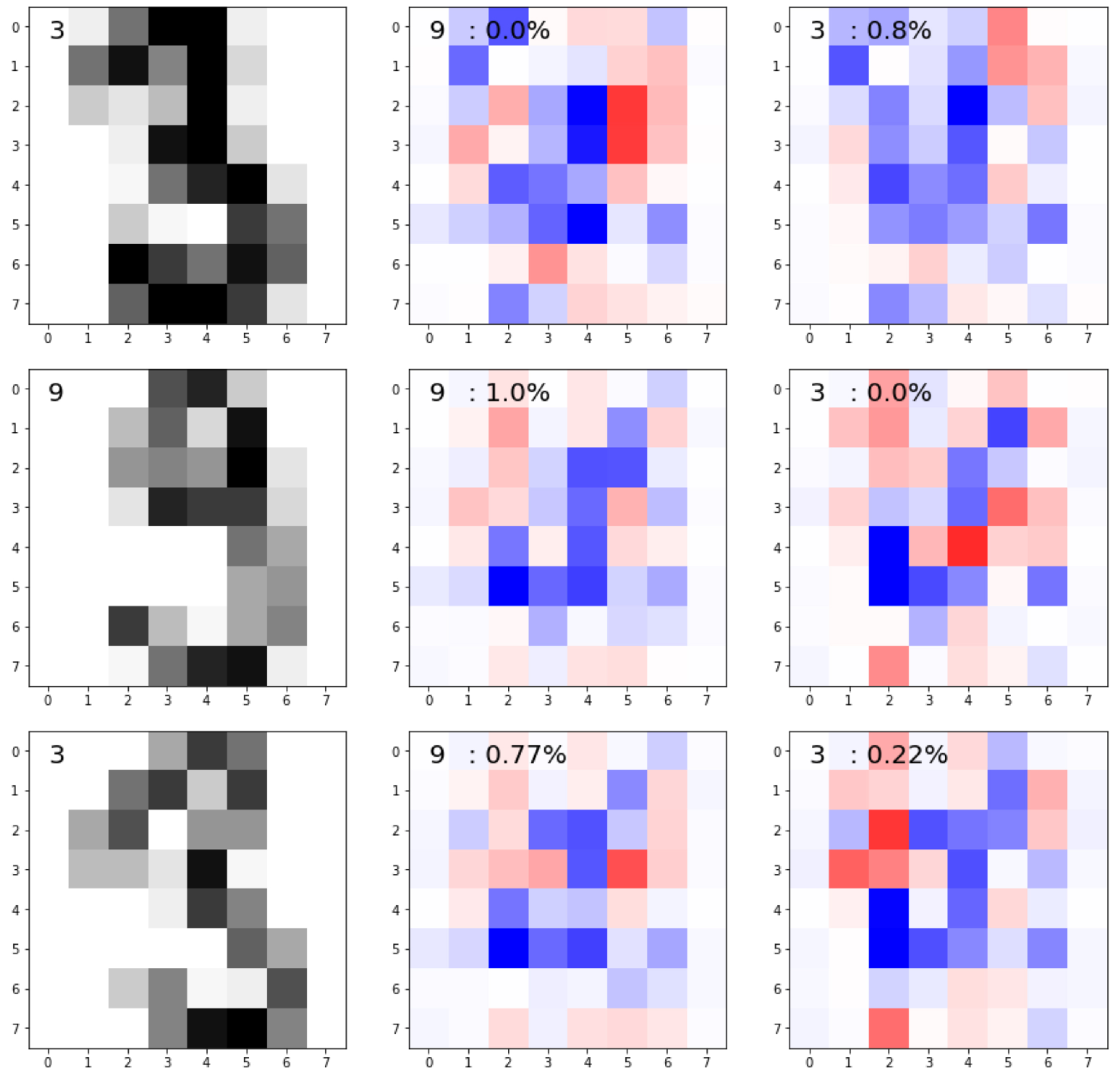}}
\caption{Example pixel contributions result. First column is the input image with the true label marked in the top-left corner. Second column is the pixel contributions (blue for positive, red for negative) for predicting digit $9$ with the probability marked also in the top-left corner. Third column is the pixel contributions for predicting digit $3$.}
\label{contributions}
\end{center}
\vskip -0.2in
\end{figure}

Figure~\ref{contributions} depicts an example of pixel contributions for three different inputs. Note that the pixel values have been scaled to cover the range $[10^{-6}, 1]$. We have reserved the $0$ value to accommodate missing pixels, whose value we do not know. This makes sense, as white pixels even though their value is $0$ in the original dataset, they do contain salient information and one expects that their contribution to be non-zero as we see in the figure. In the first two rows, model predictions match the true labels, while in the last row the model mislabels the input image.

While all the pixels provide various contributions, we focus on the most negative contributions to explain the differences between $9$ and $3$. In the first row, the negative contributions seen in the prediction of $9$ in positions $(2,5)$ and $(3,5)$ are expected as the pixels in these positions are very light because of the curvature of $3$, whereas for $9$ we expect them to be darker - see the input image in the second row. As a result they negatively contribute to the prediction of $9$.

In the second row, the negative contribution seen in the prediction of $3$ in the position $(4,4)$ is also expected as the pixel in this position is expected to be lighter due to the hollowness of $9$ in that region, whereas for $3$ we expect it to be darker as it is the inflection point for $3$ - see the input image in the third row. Finally, in the third row, the negative contribution in predicting $9$ in position $(3,5)$ is not sufficient to overcome the positive contributions that are very similar with predicting $3$. 

\section{Conclusions}
\label{conclusions}

In this study a new architecture, TableGraphNet, is proposed to jointly provide predictions and explanations in the form of feature attribution for tabular data. The design of the architecture embraces an axiomatic view of feature attribution. TableGraphNet incorporates the local accuracy and missingness property as required by Shapely values. In addition, we empirically observe that the use of the proposed augmented training strategy that exploits the missingness property, also increases the chances to satisfy the consistency property. The byproduct of adopting this axiomatic view is that TableGraphNet does not require imputation as it inherently accommodates missing data and it also obeys the symmetry-preserving property, namely the output is not impacted by the order of the attributes. Since TableGraphNet constructs a fully connected graph where each node represents an attribute, an obvious limitation is the number of attributes that can be used during training. This limitation can be overcome by more efficient implementations. Future research directions include a more thorough study of the node centric features and their impact on consistency. 

\section*{Software and Data}
\label{software}

The manuscript is accompanied by a supplementary report that further details the results obtained for each trial run, as well as the source code that will also be made publicly available with the publication of this manuscript.



\bibliography{paper}
\bibliographystyle{icml2020}

\end{document}